\documentclass[sigconf, nonacm]{acmart}

\usepackage{amsmath,amsfonts}
\usepackage{algorithm}
\usepackage{algorithmic}
\usepackage{graphicx}
\usepackage{textcomp}
\usepackage{xcolor}
\usepackage{verbatim}
\usepackage{booktabs}
\usepackage{url}

\acmConference[Conference'YY]{ACM Conference Name}{Month DD--DD, YYYY}{City, Country}
\acmYear{2026}
\copyrightyear{2026}
\acmISBN{978-x-xxxx-xxxx-x/YY/MM}
\acmDOI{10.1145/nnnnnnn.nnnnnnn}

\begin{document}

\title{Agentic Analysis for Agentic Infrastructure: An LLM-Powered Pipeline for Comparative Governance of DAO and Corporate AI Protocols}

\author{Yutian Wang}
\affiliation{
  \institution{Duke Kunshan University}
  \city{Kunshan}
  \country{China}
}
\authornotemark[2]
\author{Luyao Zhang}
\affiliation{
  \institution{Duke Kunshan University}
  \city{Kunshan}
  \country{China}
}

\authornote{The corresponding author: Luyao Zhang (email: lz183@duke.edu, address: Duke Kunshan University, No.8 Duke Ave. Kunshan, Jiangsu 215316, China.) }
\authornote{\textbf{Acknowledgments}:Yutian Wang gratefully acknowledges the Summer Research Scholar Program supervised by Professor Zhang at Duke Kunshan University.}

\begin{abstract}
As AI agent protocols proliferate, the governance structures shaping their interoperability standards remain empirically underexamined. We introduce an LLM-powered comparative pipeline for large-scale governance discourse analysis, integrating automated annotation, neural topic modeling, and multi-layer network analysis to study socio-technical power structures at scale. We validate it on two contrasting standards for agent interoperability: ERC-8004 (permissionless, on-chain) and Google A2A (corporate-led). Analyzing 4,323 governance participation records, we combine LLM-assisted coding, topic modeling, and multi-layer network analysis to examine how institutional design shapes thematic priorities and community structure. We find that while governance form influences substantive focus, both regimes exhibit comparable levels of participation inequality and community fragmentation. Discourse alignment is denser in the permissionless setting, suggesting that open governance may foster greater thematic convergence despite decentralized participation. These findings illustrate how LLM-assisted methods can advance the empirical study of technology governance, with implications for designing more equitable agentic AI standards. All data and code are openly available.
\end{abstract}

\keywords{DAO governance, consortium governance, AI agents, LLM research pipelines, blockchain}

\maketitle


\section{Introduction}


Interoperability is an inherited characteristic of decentralized applications~\cite{harvey2024}. As AI agent protocols proliferate, the governance of their interoperability standards remains empirically underexamined. Who controls the rules by which autonomous agents discover, negotiate, and coordinate across organizational boundaries? This question sits at the intersection of artificial intelligence and institutional design~\cite{evans2026,wef2026}, yet the public discourse through which such standards are negotiated has been difficult to study at scale. Prior work relies predominantly on manual coding with fixed categories, limiting both thematic discovery and structural analysis across large text corpora.


We address this gap with an LLM-powered comparative pipeline for large-scale governance discourse analysis, integrating automated annotation, neural topic modeling, and multi-layer network analysis. We validate the pipeline on two standards: ERC-8004 and Google A2A. The two protocols address the same technical problem of AI agentic communication across systems, and their governance processes are both public on GitHub or Forums. Therefore, comparing them isolates the effect of governance form on who participates and what they discuss. 

ERC-8004, which belongs to Ethereum Improvement Proposals (EIP), defines a smart-contract interface standard for on-chain AI agent identity. It ``proposes to use blockchains to discover, choose, and interact with agents across organizational boundaries without pre-existing trust''~\cite{erc8004-forum}. Its lifecycle follows the stages defined in EIP-1, and is advanced by rough consensus among forum participants and EIP editors, with no formal votes and permissionless implementations~\cite{eip1}. Google A2A introduces ``An open protocol enabling communication and interoperability between opaque agentic applications''~\cite{a2a}. It was initiated by Google and donated to the Linux Foundation in June 2025, now governed by an eight-seat Technical Steering Committee (TSC) composed entirely of corporate representatives~\cite{a2a-governance,google-donation}.


Comparing the two cases, we ask:

\textbf{RQ: Compared to corporate hierarchy, does the governance structure of a permissionless DAO really achieve a higher degree of decentralization?}

We decompose this into three sub-questions:

\begin{itemize}
  \item \textbf{RQ1 (Decision Architecture):} How do the formal decision procedures, entry rights, and authority structures of the two regimes differ?

  \item \textbf{RQ2 (Discourse Composition):} How does governance form shape the topical and argumentative composition of participation discourse?

  \item \textbf{RQ3 (Relational Networks):} How does governance form shape co-participation ties, discourse-level consensus and conflict structures, and actor--topic divisions of collaborative labor?
\end{itemize}


To answer these questions, we first reconstruct the two governance mechanisms from their official documents and visualize them as decision-flow diagrams. We then collect 4{,}323 governance participation records from their public repositories and apply an LLM-assisted annotation pipeline to label each record's argumentative function, stance, and stakeholder affiliation~\cite{carlson2026}. On top of this annotation, we run two topic-discovery methods: unsupervised BERTopic~\cite{bertopic_2022}, and LLM-inductive Thematic-LM~\cite{thematic_lm_2025}. Moreover, we run three network analyses: co-participation social network analysis (SNA)~\cite{ao2023}, discourse network analysis~\cite{leifeld2013}, and socio-semantic bipartite networks~\cite{roth2010}. The methodological triad mirrors the three sub-questions.


Three findings emerge. First, the two cases instantiate structurally opposite decision architectures: ERC-8004 advances by rough consensus with permissionless deployment, while A2A vests binding authority in a corporate technical steering committee. Second, governance form shapes the thematic focus of deliberation: DAO governance concentrates discourse on security mechanisms and protocol principles, whereas corporate governance distributes deliberation across engineering-execution workstreams. Third, both networks exhibit comparable levels of participation inequality and community fragmentation; discourse congruence is nonetheless denser in ERC-8004, reflecting tighter within-community consensus formation. 


This paper makes three contributions. Methodologically, the architecture can generalize to any large-scale governance discourse analysis, offering a computational lens on the black box of public text negotiation, communication and coordination. Empirically, it provides the first matched-case, multi-method comparison of DAO and corporate governance of agentic standardization with text corpora at scale. Theoretically, it pioneers the fundamental problem of who controls the future AI agentic infrastructure, with the interdisciplinary standing of institutional design and AI.


\section{Related Work}

Three streams of literature converge on this study. For a more specific comparison, see Appendix~\ref{app:table}.

\textbf{Decentralized governance via blockchain.} First, Harvey provides a comprehensive overview of the technologies, theories, and practices inherent to decentralized finance (DeFi) and blockchain technology~\cite{harvey2021}. Beck et al.~\cite{beck2018} provided the foundational Information System framework mapping blockchain governance along decision rights, accountability and incentives dimensions. Ziolkowski et al. ~\cite{ziolkowski2020} identified six major governance challenges specific to the blockchain system, and half of them are unique to this technology. Kiayias and Lazos~\cite{kiayias2022} systematised the field with a SoK review of blockchain governance mechanisms. More recently, Ellinger et al.~\cite{ellinger2023} explored the balanced polycentric governance of digital commons. Reineke et al.~\cite{reineke2025} developed an integrative theoretical framework, showing the evolution of decentralization concepts. Sunyaev et al.~\cite{sunyaev2026} called for purposeful, design-oriented decentralization rather than ending at ideology. Motea and Oba~\cite{motea2026} interrogated the democratic legitimacy of blockchain governance structures. 


\textbf{Governance of decentralized versus corporate structure.} Murray et al.~\cite{murray2021} examined how smart contracts and DAOs alter agency costs in corporate contracting, providing leading opinions. Lumineau et al.~\cite{lumineau2021} highlights tacitness of transactions and social implications of this technology. Rahman et al.~\cite{rahman2024} and Hunt et al.~\cite{hunt2025} analyzed the dynamics and power accumulation of platforms. Hui and Tucker~\cite{hui2025} address AI along with a decentralized ecosystem, proposing an innovative governance framework. However, these studies rely on theoretical or interview-based methods; none of them use governance-participation data or computational methods to test structural difference between DAO and corporate governance in the same domain. 


\textbf{Computational studies of cooperative work.} Researchers have long examined coordination in open, online communities. Mockus et al.~\cite{mockus2002} documented participation inequality in Apache and Mozilla. Im et al.~\cite{im2018} analyzed Wikipedia's Requests for Comments (RfC), a deliberation mechanism structurally analogous to EIP rough consensus, and revealed a persistent imbalance between deliberation and resolution. Germonprez et al.~\cite{germonprez2018} catalogued structural realities of contemporary open-source projects, including pervasive corporate engagement; Li et al.~\cite{i2021} examined code-of-conduct conversations on GitHub as a window into informal governance norms in open-source repositories. Kulakowski and Frasincar~\cite{kulakowski2023} introduced CryptoBERT, a domain-adapted BERT variant pre-trained on 3.2~million cryptocurrency social-media posts, establishing a specialized embedding backbone for blockchain-native corpora. More recently, Wu et al.~\cite{wu2024} applied sentiment and discourse analysis to six DAO forums; Stine and Agarwal~\cite{stine_agarwal_2020} proposed comparative discourse analysis via topic models; Qiao et al.~\cite{thematic_lm_2025} introduced Thematic-LM for LLM-assisted inductive thematic analysis of large corpora; Ao et al.~\cite{ao2023} used social network analysis on on-chain Aave data to show voting-power concentration; Leifeld~\cite{leifeld2013} proposed discourse network analysis concentrating on participants' stance; Roth and Cointet~\cite{roth2010} connects semantic analysis with social topology. Wang el al. ~\cite{understanding2025} found the threat of power concentration via scaled empirical analysis of scale \textit{SnapShot} data. \"{O}zdemir S\"{o}nmez et al.~\cite{ozdemir2025} quantified voting power concentration and participation apathy across DAO governance models, especially in token-based organizations. Chen et al.~\cite{chenfang2025} applied a quasi-experimental PSM-DID design to 98{,}000 Steemit users, finding that governance token ownership boosts curation effort but reduces creation novelty.


\section{Methodology}

We selected MiniMax-M2.5~\cite{minimax_m25} as the LLM backbone for its reasoning capability and low cost. The model is trained on complex real-world environments and achieves 80.2\% on SWE-Bench Verified; frontier models of this capability tier have been shown to complete complex multi-turn analytical tasks with high reliability~\cite{agentboard2024}, making them suitable for assigning nuanced governance-process labels such as \textit{Argument Type} and \textit{Consensus Signal}. For design of the comparative case, see Appendix~\ref{app:compa-design}.


\subsection{Data Collection and LLM Annotation}

We collect governance participation records from the public repositories. For ERC-8004, data originate from two sources: (1)~113 posts from the Ethereum Magician forum thread~\cite{erc8004-forum}; and (2)~36 GitHub records from the nine pull requests that directly modified \texttt{ERCS/erc-8004.md}~\cite{erc-github}. For Google A2A, data originate from three streams in the \texttt{a2aproject/A2A} repository~\cite{a2a}: (1)~3,104 issue and issue-comment records, (2)~1,955 pull requests and review-comment records, and (3)~822 GitHub Discussion records.

The raw records totaled 6,030. We removed the following data: records with fewer than 20 characters of body text, which primarily are CI notifications, merge-conflict markers, and bot-generated status messages; and records attributed to verified bot accounts. After such filtering, 4,323 records are retained (ERC-8004: 142; Google A2A: 4,181). SHA-256 checksums for all raw data fields are provided in the authors' GitHub repository~\footnote{https://github.com/kl41r3/erc8004-a2a-case-study}. See also Appendix~\ref{app:checksums}. 


The retained records was annotated with four categorical fields:

\begin{itemize}
  \item \textbf{Stakeholder Institution}: Google / MetaMask / Ethereum Foundation / Coinbase / Independent / Unknown
  \item \textbf{Argument Type}: Technical / Governance-Principle / Economic / Process / Off-topic
  \item \textbf{Stance}: Support / Oppose / Modify / Neutral / Off-topic
  \item \textbf{Consensus Signal}: Adopted / Rejected / Pending / N/A
\end{itemize}

The institution affiliations of the top~109 contributors were also manually reviewed. The cascade is detailed in Appendix~\ref{app:annotation}.


\subsection{Discourse Composition Analysis}

To characterize the topical and argumentative composition of the two governance discourses, we apply three methods of increasing inductive depth.

\subsubsection{Supervised argument typing} 
Treating the LLM-assigned \textit{argument\_type} as the record's communicative function, we test cross-case independence with a chi-square test~\cite{cohen1988} and within-ERC-8004 temporal change across three consecutive two-month phases spanning submission to mainnet deployment (phase boundaries in Appendix~\ref{app:phase}).

\subsubsection{BERTopic comparative discourse analysis}
Following Stine and Agarwal~\cite{stine_agarwal_2020}, we fit BERTopic~\cite{bertopic_2022} jointly on the combined corpus. Texts are embedded with \texttt{all-MiniLM-L6-v2}~\cite{reimers_gurevych_2019}, reduced via UMAP ($n_{\text{neighbors}}=15$, cosine, seed$=42$), and clustered with HDBSCAN (min cluster size $10$) into $K=19$ topics plus a noise class. For each topic $t$ and case $c$, the within-case share is 
$$p_t^{(c)} = n_t^{(c)}/n^{(c)}.$$ 
Cross-case divergence is measured by Jensen--Shannon divergence~\cite{lin_1991}: 
$$\mathrm{JSD}(p,q) = \tfrac{1}{2}\mathrm{KL}(p,m) + \tfrac{1}{2}\mathrm{KL}(q,m)$$ 
with 
$$\mathrm{KL}(p, m) = \sum_{i} p(i) \log \frac{p(i)}{m(i)}, \qquad m = \tfrac{1}{2}(p+q),$$
and $\mathrm{JSD}\!=\!0$ denotes identical distributions, $\mathrm{JSD}\!=\!1$ disjoint.

To assess whether ERC-8004's topic concentration is an artefact of a general-purpose embedding model rather than a genuine structural property, we also embed the 142 ERC-8004 records with CryptoBERT, a model further pre-trained on cryptocurrency social-media posts (StockTwits, Reddit, Telegram, Twitter)~\cite{kulakowski2023}.

\subsubsection{Thematic-LM inductive themes} To complement the embedding based view with human-interpretable labels, we apply the method of Thematic-LM~\cite{thematic_lm_2025}, a four-stage multi-agent pipeline: (1)~\emph{open coding} assigns a short code to each record; (2)~\emph{aggregation} groups 300 sampled codes into 14 raw clusters; (3)~\emph{codebook review} merges and refines to 19 themes (T01--T19); (4)~\emph{theme assignment} labels every record with its best-fitting theme, or \textit{Unclassified} when confidence is insufficient. We again report per-theme shares and JSD between cases.


\subsection{Relational Network Analysis}
\label{sec:methods-network}

We construct three complementary networks for each case, each layer adding discursive information to the previous. All graphs use actors as nodes; what changes is the semantic content of an edge.

\subsubsection{Co-participation network (SNA)}
Following Ao et al.~\cite{ao2023}, an undirected edge connects two contributors who both posted to the same discussion thread (forum topic, GitHub issue, pull request, or discussion); edge multiplicity counts co-occurrences. We report six structural measures:

\begin{itemize}
  \item Density: the realized fraction of possible ties. Computed by $$\rho = \frac{2E}{N(N-1)}$$ where $E$ denotes the number of edges and $N$ denotes the number of nodes.
  
  \item Degree Gini: inequality of interaction counts ($d_i$ is node degree). Computed by $$G = \frac{\sum_{i,j}|d_i - d_j|}{2N\sum_i d_i}$$ where $d_i$ is the number of edges of node $i$. $G = 0$ implies perfectly equal participation; $G \to 1$ implies that interactions are concentrated in a small elite.
  
  \item Components and giant-component ratio: fragmentation into disconnected threads. Computed by $$\mathrm{GCR} = N_{\max}/N.$$
  
  \item Newman--Girvan modularity: reported for institution partitions and Louvain partitions~\cite{blondel2008}~\cite{newman2006}. Computed by $$Q = \frac{1}{2m} \sum_{i,j} \left[ A_{ij} - \frac{k_i k_j}{2m} \right] \delta(c_i, c_j)$$ where $m$ is the total number of edges, $A_{ij}$ is the adjacency matrix entry, $k_i$ is the degree of node $i$, and $\delta(c_i, c_j) = 1$ if nodes $i$ and $j$ belong to the same community, 0 otherwise.
  
  \item Core--periphery: Borgatti--Everett coreness~\cite{borgatti2000}. Computed by $$\rho_{\mathrm{BE}} = \mathrm{corr}(A,\,\Delta), \qquad \Delta_{ij} = \delta_i \cdot \delta_j$$ where $\delta_i \in \{0,1\}$ is the coreness label of node $i$. Statistical significance is evaluated by comparing the observed $\rho_{\mathrm{BE}}$ against a null distribution of random graphs that preserve the empirical degree sequence (configuration model), following the $q$-$s$ test of Kojaku and Masuda~\cite{kojaku2018}.

  \item Betweenness centrality and network efficiency: identifies governance brokers---actors who mediate between otherwise disconnected groups:
  $$b(v) = \sum_{s \neq v \neq t} \frac{\sigma_{st}(v)}{\sigma_{st}}$$
  where $\sigma_{st}$ is the total number of shortest $s$--$t$ paths and $\sigma_{st}(v)$ the number that pass through $v$. 

  \item Network efficiency: mean normalized harmonic centrality, measured by
  $$\bar{h} = \frac{1}{n(n-1)}\sum_{u \neq v} d(u,v)^{-1}$$
  where $d(u,v)^{-1}=0$ for unreachable pairs~\cite{latora2001}. 
\end{itemize}


\subsubsection{Discourse Network Analysis (DNA)}

Following Leifeld~\cite{leifeld2013}, we construct the co-participation layer with stance-aware edges that capture shared discursive positions. Let $A$ be the actor set and $T=\{T_{01},\ldots,T_{19}\}$ the Thematic-LM codebook. For each record we encode stance into Support$=+1$, Modify$=+0.5$, Neutral$=0$, Oppose$=-1$ and obtain an actor--theme stance matrix $M$ similar to the pseudo-matrix below:

\begin{center}
\small
\begin{tabular}{lccc}
\toprule
       & Theme A & Theme B & Theme C \\
\midrule
Alice  & $+1$    & $-1$    & ---     \\
Bob    & $+1$    & $+0.5$  & $0$     \\
Carol  & ---     & $-1$    & $+1$    \\
\bottomrule
\end{tabular}
\end{center}

Two networks are projected from $M$: the \emph{congruence network} $G^+$ connects actors who take same-sign stances on at least one shared theme (Alice and Bob agree on Theme A), and the \emph{conflict network} $G^-$ connects actors who take strictly opposite-sign stances (Alice and Bob conflict on Theme B); edge weights count the number of themes satisfying each criterion. We report density, Louvain modularity and betweenness centrality~$b(v)$ of the top-5 discourse brokers.


\subsubsection{Socio-semantic bipartite network}
Following Roth and Cointet~\cite{roth2010}, we construct a two-mode (bipartite) network $\mathcal{B} = (A \cup T,\, E)$, where $A$ is the set of actors, $T$ the set of Thematic-LM themes, and an edge $(a, t) \in E$ exists whenever actor $a$ authored at least one record assigned to theme $t$. The weight $B_{at}$ equals the number of such records, forming a non-negative integer matrix $B \in \mathbb{Z}_{\ge 0}^{|A|\times|T|}$.

Two one-mode projections are derived from $B$. The \emph{actor--actor projection} $W^A = BB^\top$ links any two actors by the number of themes they co-discussed (self-loops removed); the \emph{theme--theme projection} $W^T = B^\top B$ links any two themes by the number of actors who engaged in both.

Per-actor topic diversity is measured by Shannon entropy:
\begin{equation*}
  H(a) = -\sum_{t=1}^{|T|} \hat{p}_{at} \log_2 \hat{p}_{at},
  \qquad \hat{p}_{at} = \frac{B_{at}}{\textstyle\sum_{t'} B_{at'},}
  \label{eq:entropy}
\end{equation*}
where $H(a)=0$ for a pure specialist (all posts in one theme) and $H(a)=\log_2|T|$ for a perfect generalist (uniform spread across all $|T|$ themes).

We further characterize the distribution of $\{H(a)\}_{a \in A}$ via its Gini coefficient. The higher the Gini, the more thematic breadth is concentrated in a few actors. We also measure per-theme actor concentration as the Gini of the column vector $\{B_{at}\}_{a \in A}$ for each~$t$. 

The thematic overlap coefficient is measured by 
\begin{equation*}
  \Omega = \frac{|T_1 \cap T_2|}{\min(|T_1|, |T_2|)},
  \label{eq:overlap}
\end{equation*}
where $T_c$ denotes the active theme set of case $c \in \{1,2\}$, quantifies cross-case thematic alignment; $\Omega = 1$ means one community's thematic space is a subset of the other's. 

Actor-set sizes across the three filtering stages are detailed in Appendix~\ref{app:actor-filter}.


\section{Results}


\subsection{Decision Architectures}

\begin{figure*}[htbp]
  \centering
  \includegraphics[width=\linewidth]{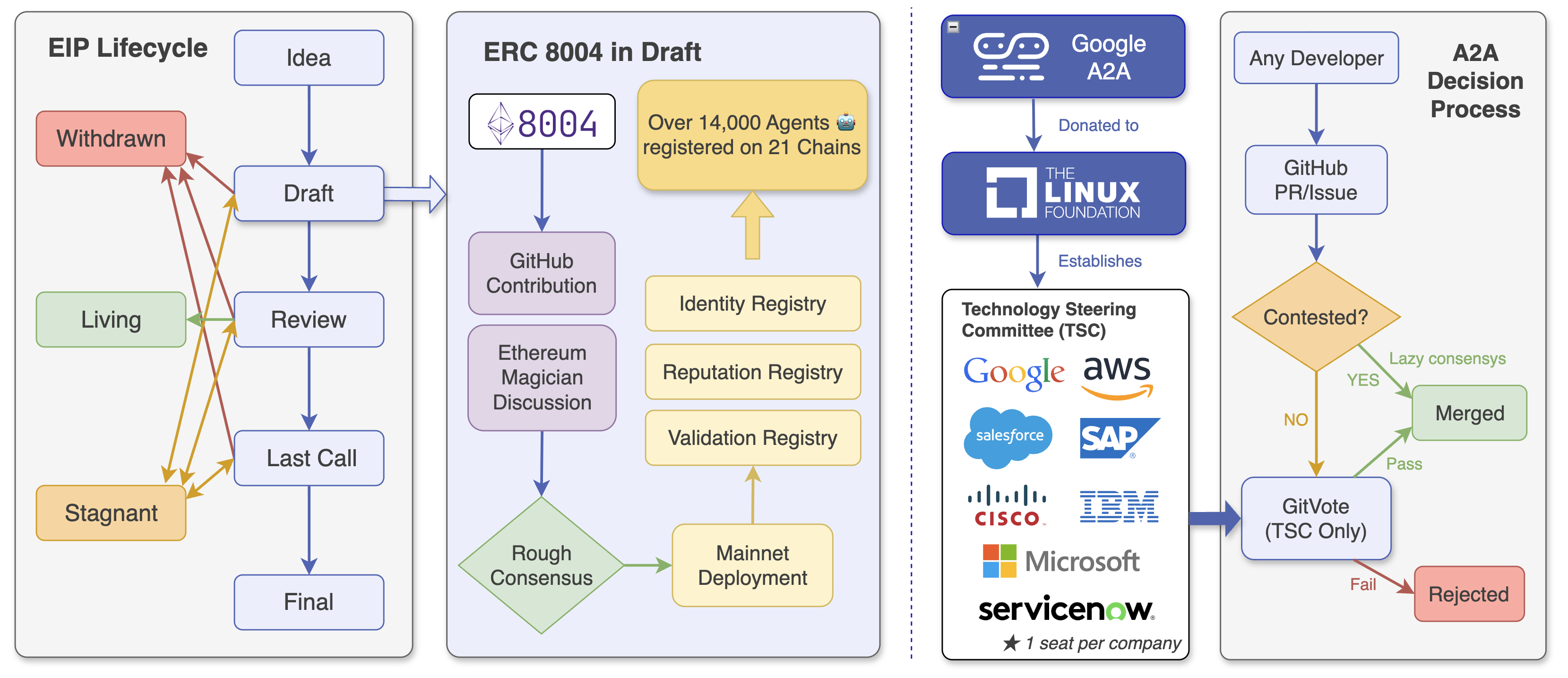}
  \caption{Left: ERC-8004 governance decision flow. Right: A2A governance structure and decision flow.}
  \Description{Two-panel flow diagram. Left panel: ERC-8004 lifecycle stages (Idea, Draft, Review, Last Call, Final/Living/Stagnant/Withdrawn) with arrows showing progression via rough consensus and EIP editor review; a dashed arrow indicates decoupled mainnet deployment. Right panel: A2A governance hierarchy showing Google/Linux Foundation stewardship, an eight-seat TSC with one seat per corporate member, and two decision paths: lazy consensus for routine PRs and GitVote for contested changes.}
  \label{fig:decision}
\end{figure*}

ERC-8004 and Google A2A represent contrasting governance arche-types, as detailed in figure~\ref{fig:decision}. ERC-8004 remains at the stage of ``draft'' when the data was fetched (Mar 2026), nevertheless, its three canonical registry contracts were deployed on Ethereum mainnet on January~29, 2026~\cite{0129}. Proposals are discussed through the Ethereum Magicians forum and are recorded on GitHub. Ownership of Google A2A transitioned from Google-controlled to Linux Foundation governance, and Contested decisions result in a GitVote~\cite{gitvote}. For more details and pseudocode, see Appendix~\ref{app:architectures}.


\subsection{Discourse Composition}


\subsubsection{Supervised argument typing}

\begin{figure}[htbp]
  \centering
  \includegraphics[width=\linewidth]{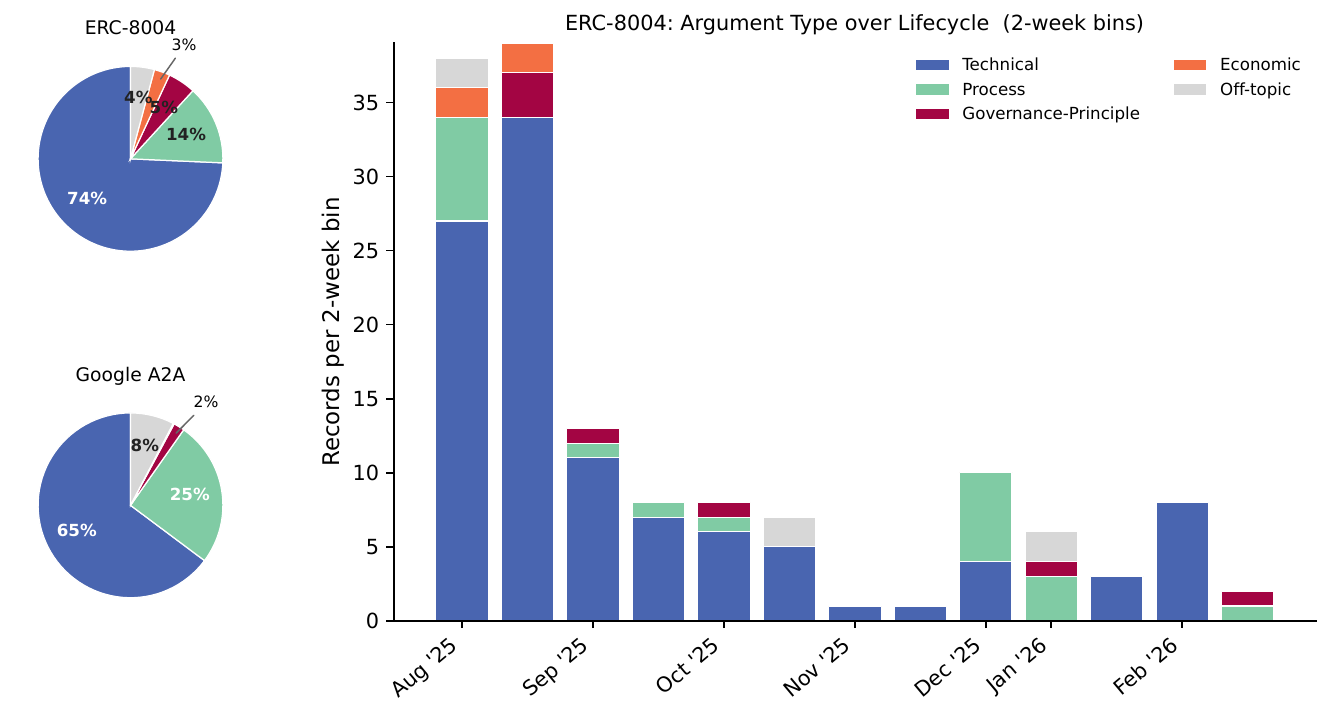}
  \caption{Left: argument-type distribution of both cases. Right: ERC-8004 argument-type distribution across three two-month phases. Technical arguments dominate in both regimes, but corporate governance carries roughly double the procedural coordination burden.}
  \Description{Two-panel figure. Left panel shows pie charts comparing argument-type categories (Technical, Process, Governance-Principle, Economic, Off-topic) between ERC-8004 and Google A2A. ERC-8004: 74.3\% Technical, 13.9\% Process, 5.4\% Governance-Principle, 1.4\% Economic, 5.0\% Off-topic. Google A2A: 62.1\% Technical, 25.4\% Process, 2.7\% Governance-Principle, 0.7\% Economic, 9.0\% Off-topic. Right panel is a stacked bar chart for ERC-8004 across Phase 1 (Aug--Oct 2025, $n=100$), Phase 2 (Oct--Dec 2025, $n=11$), and Phase 3 (Dec 2025--Feb 2026, $n=17$). Technical arguments dominate Phases 1--2 ($\geq$80\%); Process discussion surges to 53\% in Phase 3.}
  \label{fig:topic-erc}
\end{figure}

\begin{figure*}[htbp]
  \centering
  \includegraphics[width=0.9\linewidth]{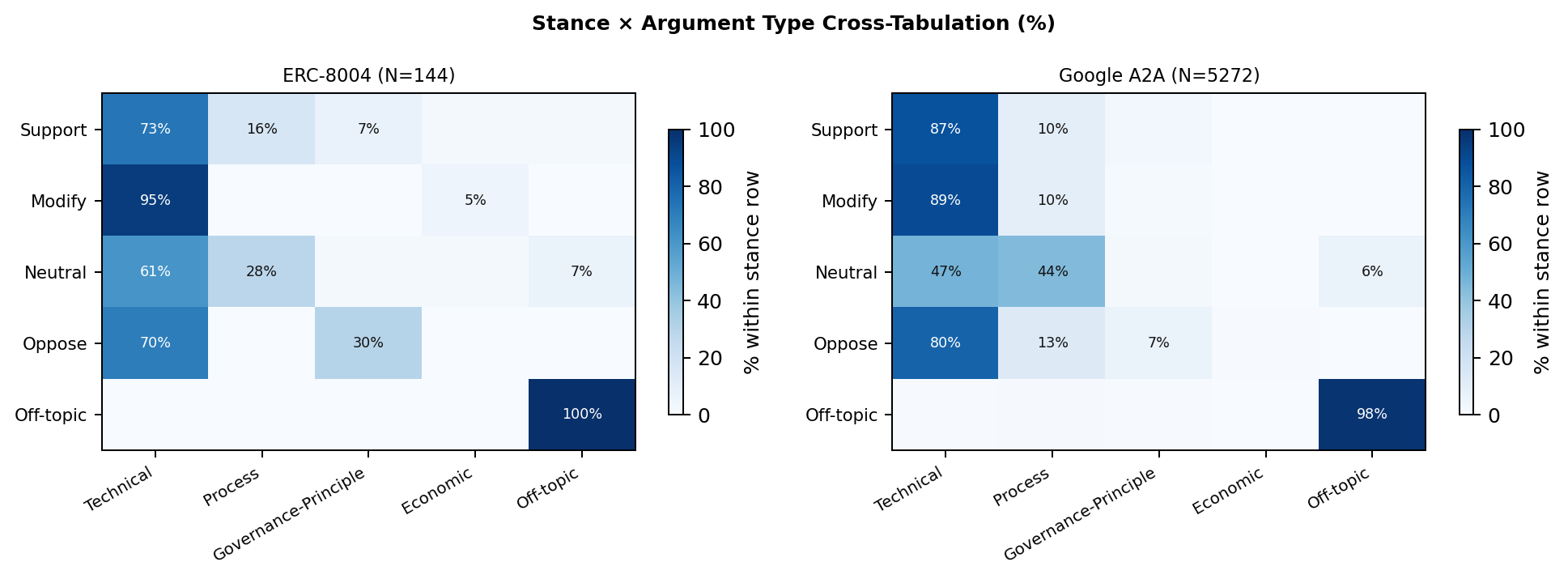}
  \caption{Stance $\times$ argument-type cross-tabulation (\% within stance row).}
  \Description{Two side-by-side heatmaps, one for ERC-8004 (left) and one for Google A2A (right). Rows represent stance categories (Support, Modify, Neutral, Oppose, Off-topic); columns represent argument types (Technical, Process, Governance-Principle, Economic, Off-topic). Cell shading encodes percentage within each stance row. Both cases are Technical-dominated under Support and Modify; the Neutral row in A2A shows a markedly higher Process share (44\%) compared to ERC-8004 (26\%), and Governance-Principle arguments are more prominent under Oppose in ERC-8004.}
  \label{fig:stance-heatmap}
\end{figure*}

Figure~\ref{fig:topic-erc} compares the LLM-assigned argument type distributions. Both communities are predominantly Technical (74.3\% in ERC-8004 vs.\ 62.1\% in A2A), but A2A devotes almost twice the share to Process arguments (25.4\% vs.\ 13.9\%). The cross-case difference is significant with a small effect size ($\chi^2(3) = 52.88$, $p < .001$, Cram\'{e}r's $V = .103$): both cases remain technically grounded, but corporate governance carries substantively heavier coordination overhead.

Within ERC-8004, argument composition shifts significantly across three two-month phases ($\chi^2(6)=25.32$, $p<.001$, $V=.315$). Technical arguments dominate Phases~1--2 ($\geq$80\%); in Phase~3 Process discussion surges to 53\% as deliberation moves from substantive design to editorial ratification. This two-stage pattern is consistent with rough-consensus norms in open standards bodies. Small-$n$ caution applies to Phases~2--3 ($n=11$ and $n=17$).

Figure~\ref{fig:stance-heatmap} cross-tabulates stance with argument type. Support and Modify stances are dominated by Technical arguments in both cases (ERC Support 73\%; A2A Support 87\%): substantive engagement is predominantly technical regardless of governance form. Governance-Principle arguments, particularly under \textit{Oppose}, are more visible in ERC-8004, reflecting the principled deliberation characteristic of the EIP process. The sharpest cross-case contrast appears in the \textit{Neutral} row: 44\% of A2A Neutral records are Process in nature, against 26\% in ERC-8004---corporate governance generates procedural commentary even among non-committal participants.


\subsubsection{BERTopic}

\begin{figure}[htbp]
  \centering
  \includegraphics[width=0.95\linewidth]{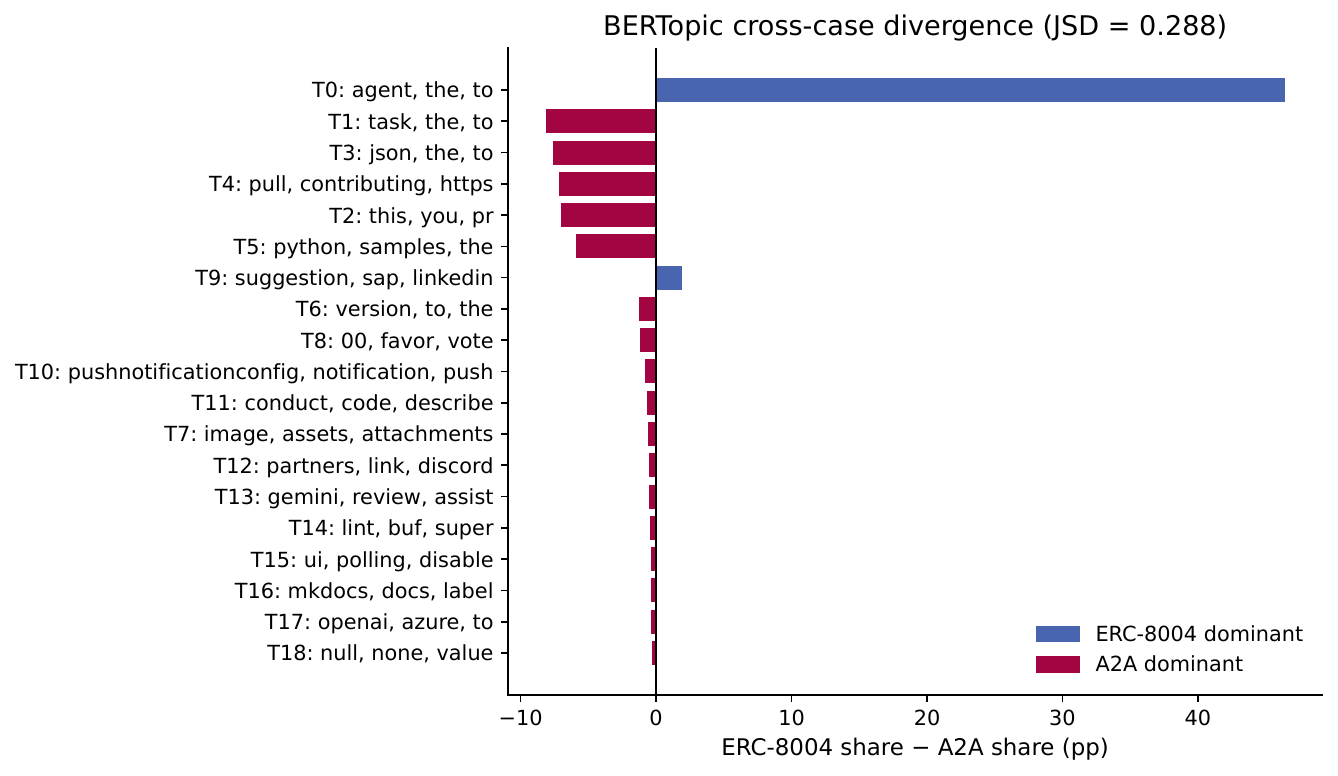}
  \caption{BERTopic cross-case divergence. ERC-8004 concentrates on foundational agent architecture, while A2A distributes deliberation across engineering-execution workstreams.}
  \Description{Horizontal diverging bar chart with 20 BERTopic topics (19 named topics plus a noise class) sorted by the absolute difference in within-case share between ERC-8004 and Google A2A. Blue bars extend left to indicate ERC-8004-dominant topics (e.g., Topic 0: broad agent discourse); orange bars extend right for A2A-dominant topics (e.g., Task/Message management, JSON/proto spec, PR-contribution workflows). The chart highlights moderate but meaningful structural separation at global JSD$=0.288$.}
  \label{fig:bertopic-divergence}
\end{figure}

BERTopic identifies 19 topics plus a noise class over the combined corpus. The global Jensen--Shannon divergence between the two cases' topic distributions is $\mathrm{JSD}_{\text{BERTopic}}=0.288$, a moderate but meaningful structural separation (Figure~\ref{fig:bertopic-divergence}). ERC-8004 is strikingly concentrated: 67.6\% of its records fall into Topic~0 (\textit{agent, agents, for, of}), the broad agent-discourse cluster, versus 21.2\% for A2A. Implementation-heavy topics---Task/Message management (8.2\%), JSON/proto spec (7.7\%), PR-contribution workflows (7.2\%), SDK samples (5.9\%)---carry substantial A2A weight but zero ERC-8004 records, confirming that low-level engineering is absent from the EIP forum. The only topic with a higher ERC share than A2A is Topic~9 (\textit{suggestion, sap, linkedin}; 2.8\% vs.\ 0.9\%), signalling corporate voices (SAP, LinkedIn) entering the EIP thread.  

\begin{figure}[htbp]
  \centering
  \includegraphics[width=0.95\linewidth]{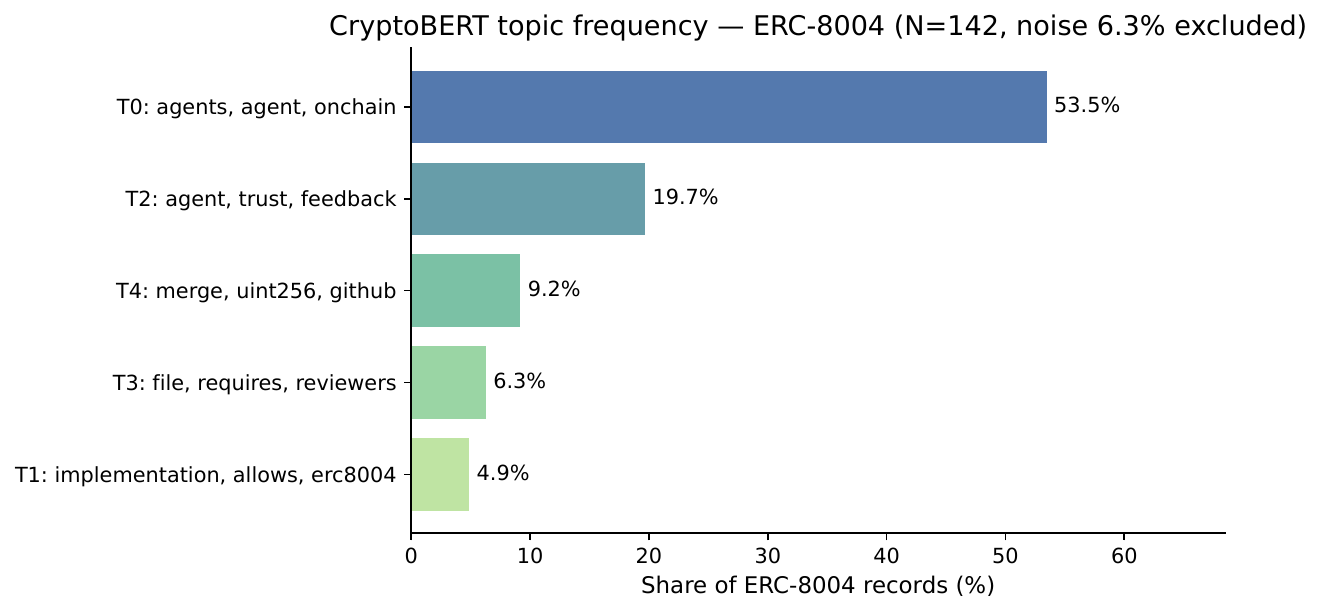}
  \caption{CryptoBERT topic frequency for ERC-8004.}
  \Description{Horizontal bar chart showing CryptoBERT topic frequency distribution for ERC-8004 records.}
  \label{fig:cryptobert-frequency}
\end{figure}

CryptoBERT on ERC data shows more precise results. It divides the "agent" topic to T0 (Onchain, Reputation) and T2 (Trust, Feedback), and the total percentage is 73.2\%, which aligns with the 67.6\% of pure BERTopic (figure~\ref{fig:cryptobert-frequency}).

\subsubsection{Thematic-LM}
\begin{figure}[htbp]
  \centering
  \includegraphics[width=\linewidth]{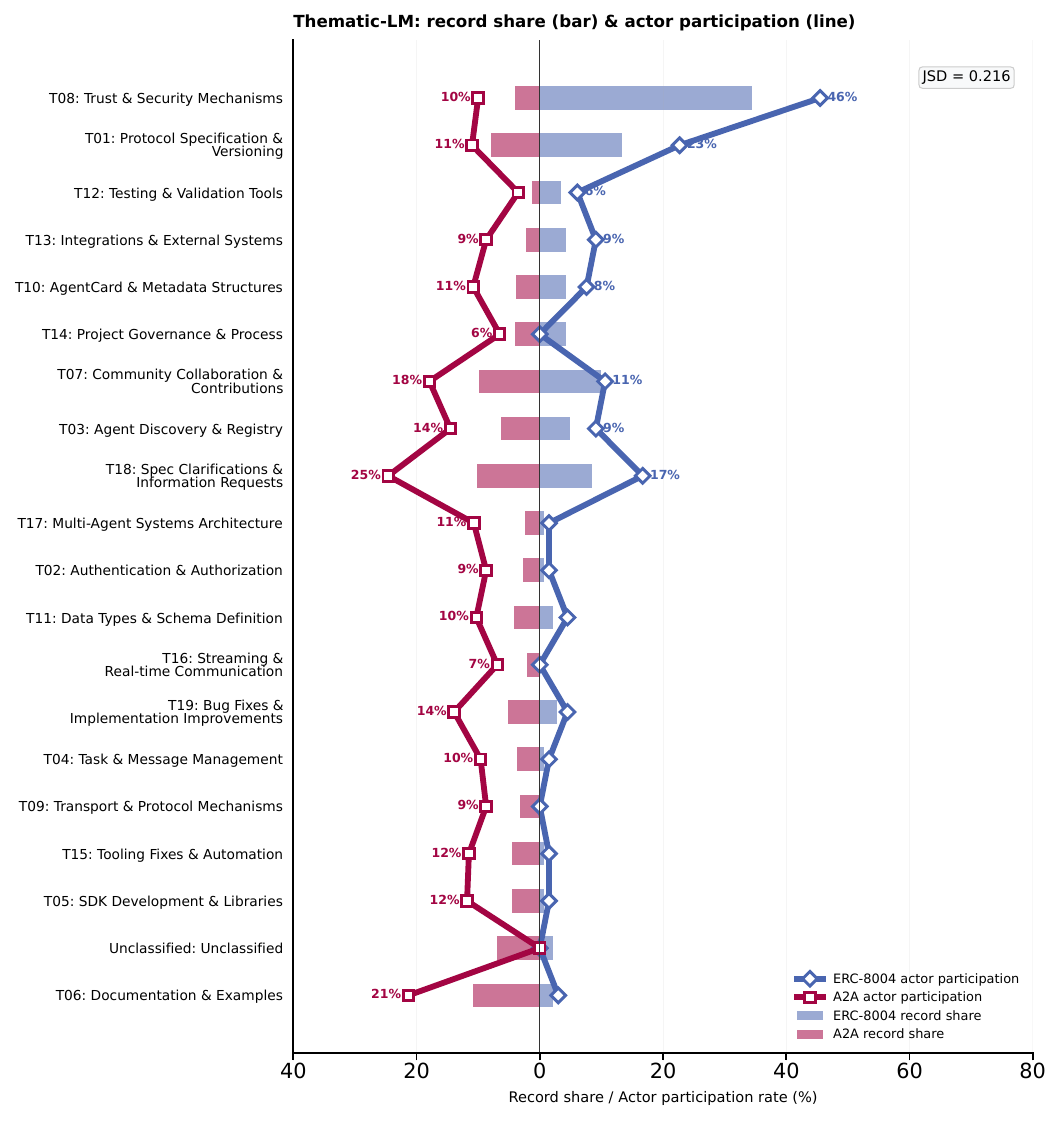}
  \caption{Thematic-LM overlaid chart. Bars show per-case record share; lines show actor participation rate. The DAO concentrates deliberation on security and trust; the corporate project distributes effort across documentation, tooling, and engineering-execution themes.}
  \Description{Overlaid diverging bar and line chart. Bars show record share diverging from center: ERC-8004 bars extend right, A2A bars extend left. Lines with diamond (ERC) and square (A2A) markers show actor participation rates, overlaid on the same axes. Theme rows sorted by ERC minus A2A share difference. Nineteen themes with full codebook labels on the y-axis.}
  \label{fig:combined-themes}
\end{figure}

The Thematic-LM pipeline yields a 19-theme codebook (T01--T19). Figure~\ref{fig:combined-themes} overlays per-case record share (diverging bars) with per-theme actor participation rate (lines with markers). Table~\ref{tab:themes-compact} reports the six themes with the largest cross-case divergence. The global Jensen--Shannon divergence is $\mathrm{JSD}=0.216$.

The most discriminating theme is \textbf{T08 Trust \& Security Mechanisms}. It covers agent-trust scoring, reputation schemes and on-chain credential verification---the foundational problem of establishing trustworthy agency. By contrast, A2A spreads deliberation across T06 Documentation \& Examples, T18 Clarifications \& Information Requests, T07 Community Collaboration \& Contributions, and T01 Protocol Specification \& Versioning. A2A additionally covers three themes: T09 (Transport \& Protocol Mechanisms), T14 (Project Governance \& Process), and T16 (Streaming \& Real-time Communication). T09 and T16 are purely engineering-execution concerns entirely absent from the EIP forum; T14's ERC-8004 records were authored by automated review bots and thus fall outside the actor network.

The two methods agree on the direction and approximate magnitude of divergence, with BERTopic's $\mathrm{JSD}=0.288$ slightly exceeding Thematic-LM's $\mathrm{JSD}=0.216$. The gap is expected: Thematic-LM compresses divergence by absorbing related concerns into shared conceptual themes, while BERTopic retains finer-grained lexical distinctions (ERC's ``trustless'', ``reputation'' versus A2A's ``samples'', ``proto''). Together the two estimates bracket the structural divergence and confirm its robustness: the two governance processes occupy meaningfully different but overlapping discourse spaces, with ERC-8004 concentrated on the constitutive layer of the design space (``what to build and why'') and A2A on executive ones (how to build, document, and ship it).

\begin{table}[htbp]
  \caption{Thematic-LM themes with largest cross-case divergence. The result is aligned with figure~\ref{fig:bertopic-divergence}.}
  \label{tab:themes-compact}
  \centering
  \small
  \begin{tabular}{llrrr}
    \hline
    \textbf{ID} & \textbf{Theme} & \textbf{ERC \%} & \textbf{A2A \%} & $\Delta$ \\
    \hline
    T08 & Trust \& Security        & \textbf{34.5} & 4.0  & $+30.5$ \\
    T01 & Protocol Specification   & \textbf{13.4} & 7.9  & $+5.5$  \\
    T06 & Documentation \& Examples& 2.1  & \textbf{10.8} & $-8.7$  \\
    T05 & SDK Development          & 0.7  & \textbf{4.6}  & $-3.9$  \\
    T15 & Tooling \& Automation    & 0.7  & \textbf{4.5}  & $-3.8$  \\
    T09 & Transport Mechanisms     & 0.0  & \textbf{3.2}  & $-3.2$  \\
    \hline
  \end{tabular}
\end{table}

\subsection{Relational Networks}


\subsubsection{Co-participation network (SNA): core and periphery}
\begin{figure}[t]
  \centering
  \includegraphics[width=\linewidth]{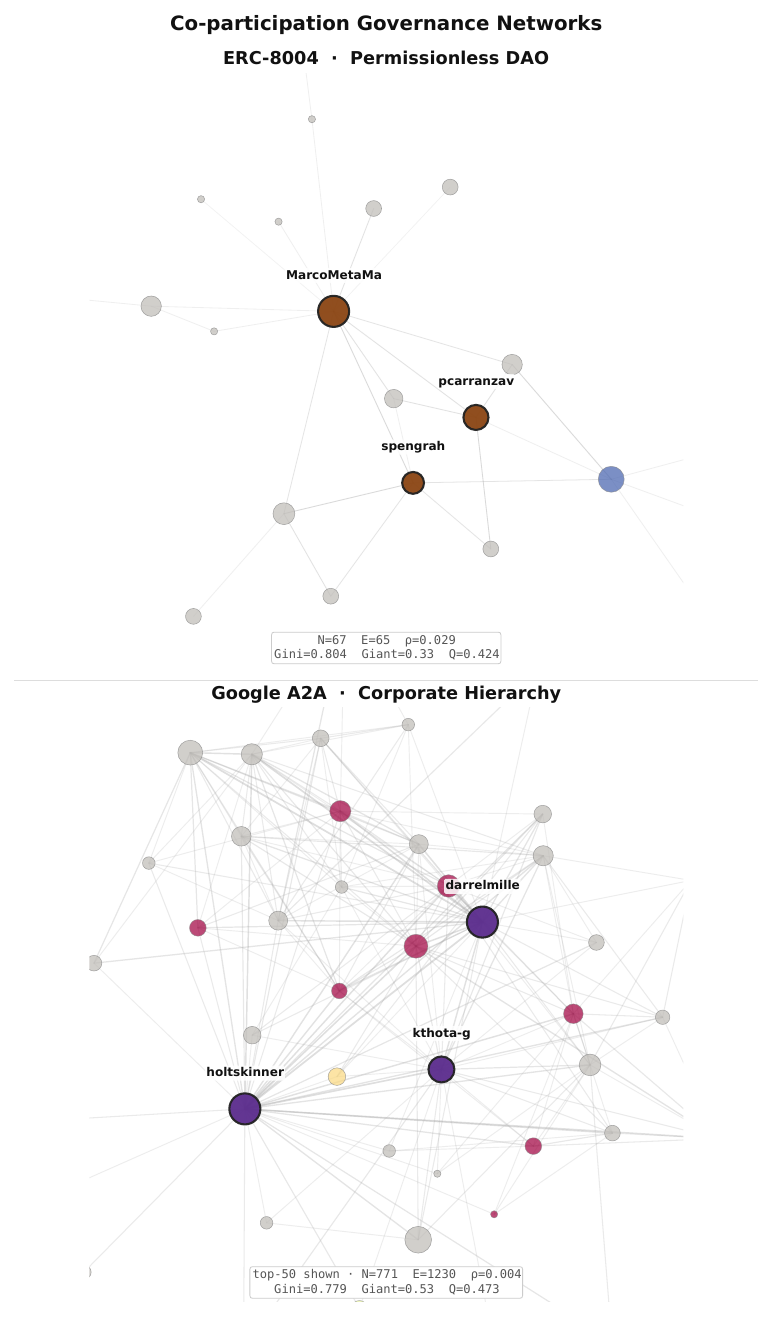}
  \caption{Your network diagram caption here.}
  \label{app:network}
\end{figure}

Table~\ref{tab:sna} reports structural metrics for the co-participation networks (ERC-8004: $N=67$; Google A2A: $N=771$). For network visualization, see~\ref{app:network}.

Both communities exhibit a high degree of inequality. The Gini coefficient of degree is 0.804 for ERC-8004 and 0.779 for A2A, indicating that in both cases, a small elite drives the majority of interactions. The top-3 contributors account for 32.3\% of all ERC-8004 interactions and 14.9\% of A2A interactions. In A2A, two out of three of Google employees, and another is from Microsoft. In ERC-8004, the three contributors are: MarcoMetaMask (Marco De Rossi), who obviously by name is from \textit{Metamask}; spengrah (Spencer Graham), cofounder of \textit{Hats Protocol}; and pcarranzav (Pablo Carranza V\'elez), who is from \textit{The Graph}, a decentralized protocol for indexing and querying blockchain data. Their formal organizational mandate is indeterminate from public records, but their persistent participation accrues reputational standing. 

Both networks are structurally fragmented, and neither shows statistically significant core-periphery structure. Louvain community detection recovers 46 communities for ERC-8004 and 358 for A2A, closely matching the component counts and confirming that participation organizes around parallel threads rather than a coherent deliberative body. Institution labels also do not predict interaction structure in either case. The giant component covers 32.8\% of ERC-8004 nodes and 53.4\% of A2A nodes; 333 of 771 A2A participants (43.2\%) are complete isolates with no co-participation edges. Density indicates that the vast majority of possible co-participation pairs are unrealized. Under the Borgatti-Everett test~\cite{borgatti2000}, ERC-8004 returns $p = .095$ and A2A returns $p = 1.000$; neither crosses the $\alpha = .05$ threshold.

\begin{table}[htbp]
  \caption{Governance Network Structural Metrics. Despite structurally opposite decision architectures, both networks exhibit comparably high participation inequality: a small elite drives most interactions regardless of governance form.}
  \label{tab:sna}
  \centering
  \begin{tabular}{lrr}
    \hline
    \textbf{Metric} & \textbf{ERC-8004} & \textbf{Google A2A} \\
    \hline
    Nodes                       & 67       & 771     \\
    Edges                       & 65       & 1,230   \\
    Density                     & 0.029    & 0.004   \\
    Gini (degree)               & 0.804    & 0.779   \\
    Top-3 degree share          & 32.3\%   & 14.9\%  \\
    Components                  & 43       & 346     \\
    Giant component ratio       & 0.328    & 0.534   \\
    Modularity---institution    & $-$0.059 & $-$0.034 \\
    Modularity---Louvain        & 0.425    & 0.473   \\
    Louvain communities         & 46       & 358     \\
    CP $p$-value (BE test)      & 0.095    & 1.000   \\
    CP significant ($p<.05$)    & No       & No      \\
    Top-1 betweenness centrality & 0.069    & 0.136   \\
    Betweenness Gini            & 0.931    & 0.979   \\
    Top-3 betweenness share     & 70.8\%   & 48.5\%  \\
    Network efficiency $\bar{h}$& 0.050    & 0.110   \\
    \hline
  \end{tabular}
\end{table}


\subsubsection{Discourse network: congruence and conflict}
The stance layer (Table~\ref{tab:dna}) reveals structures that are invisible in the co-participation graph.

ERC-8004 achieves denser within-community consensus, while A2A generates far greater absolute conflict volume. Congruence density is $0.148$ for ERC-8004 versus $0.082$ for A2A: within the tighter EIP community, participants more often share positions. Conflict edges in A2A are 34$\times$ larger in absolute count (2{,}531 vs.\ 74), which is not only an effect of actor numbers, but also from the broader range of technical positions supported by a multi-vendor engineering project.

The top-3 betweenness share in the congruence network drops to $34.5\%$ for ERC-8004 and $12.2\%$ for A2A, compared with $70.8\%$ and $48.5\%$ in the co-participation network (Table~\ref{tab:sna}). Discourse brokerage is more distributed than structural interaction brokerage in both cases; yet ERC-8004's congruence network remains markedly more concentrated, consistent with a compact EIP core mediating contested positions on trust and protocol security.

\begin{table}[htbp]
  \caption{Discourse Network Analysis metrics. The DAO achieves denser within-community agreement; the corporate regime generates a much larger volume of conflict edges, reflecting the broader technical surface of a multi-vendor engineering project.}
  \label{tab:dna}
  \centering
  \small
  \begin{tabular}{lrr}
    \hline
    \textbf{Metric} & \textbf{ERC-8004} & \textbf{Google A2A} \\
    \hline
    Actors                              & 66      & 710      \\
    Active themes                       & 17 / 19 & 19 / 19  \\
    Congruence edges                    & 318     & 20{,}638 \\
    Congruence density                  & 0.148   & 0.082    \\
    Congruence modularity (Louvain)     & 0.2886  & 0.2453   \\
    Top-1 betweenness centrality        & 0.061   & 0.018    \\
    Betweenness Gini                    & 0.859   & 0.889    \\
    Top-3 betweenness share             & 34.5\%  & 12.2\%   \\
    Conflict edges                      & 74      & 2{,}531  \\
    Mean actor theme diversity          & 1.470    & 2.211   \\
    \hline
  \end{tabular}
\end{table}


\subsubsection{Socio-semantic bipartite network: thematic concentration of discursive labor}
\begin{figure}[htbp]
  \centering
  \includegraphics[width=0.95\linewidth]{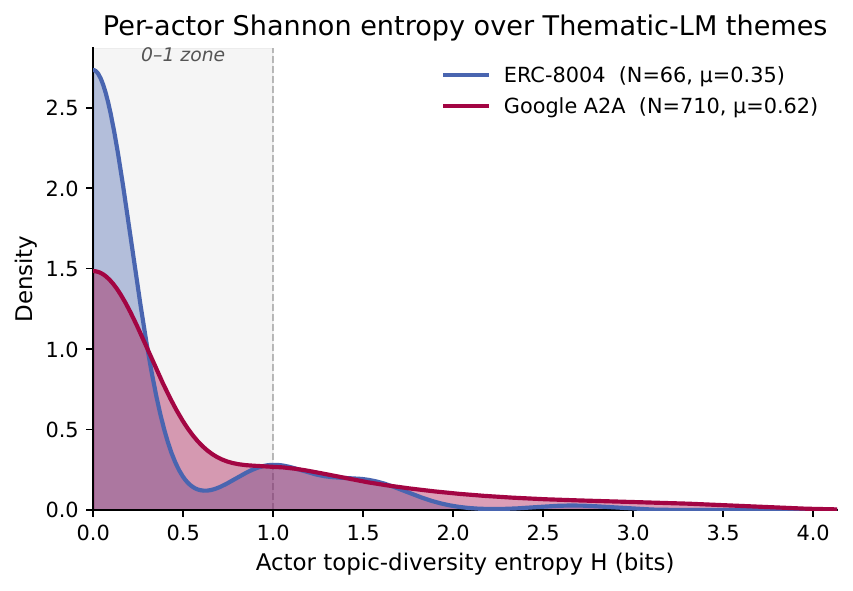}
  \caption{Per-actor Shannon entropy over Thematic-LM themes. $H=0$ marks pure specialists; the $H=0$ bin dominates both distributions.}
  \Description{Density curve plot showing the distribution of per-actor Shannon entropy $H$ over Thematic-LM themes for ERC-8004 (blue, $N=66$) and Google A2A (red, $N=710$). Both distributions are strongly right-skewed with a dominant spike at $H=0$ (pure specialists who engaged only a single theme). The A2A distribution has a heavier tail toward higher entropy values, consistent with its higher mean $H=0.617$ versus ERC-8004's mean $H=0.348$, reflecting that A2A core contributors range across more themes on average.}
  \label{fig:ss-entropy}
\end{figure}

The actor--theme bipartite layer (Table~\ref{tab:ss}, Figure~\ref{fig:ss-entropy}) reveals where each governance community directs its deliberative effort. Peripheral participation is the structural baseline in both cases: median actor Shannon entropy $H = 0$ in both ERC-8004 and A2A, meaning the majority of contributors engage a single theme across their entire record~\cite{mockus2002}. This is not a governance-specific feature but an intrinsic property of large-scale deliberative systems.

\begin{table}[htbp]
  \caption{Socio-semantic bipartite network metrics. Both regimes share the same thematic space, but governance form drives which themes absorb deliberative effort and how tightly specialized those subgroups become.}
  \label{tab:ss}
  \centering
  \small
  \begin{tabular}{lrr}
    \hline
    \textbf{Metric} & \textbf{ERC-8004} & \textbf{Google A2A} \\
    \hline
    Actors                              & 66     & 710     \\
    Active themes                       & 17/19  & 19/19   \\
    Actor--actor projection edges       & 645    & 59{,}007\\
    Mean actor entropy $H$              & 0.348  & 0.617   \\
    Median actor entropy $H$            & 0      & 0       \\
    Gini$(H)$                           & 0.773  & 0.707   \\
    Max $H$                             & 2.664  & 3.834   \\
    Mean theme Gini (actor concentration)& 0.085 & 0.453   \\
    Most-discussed theme                & T08    & T06     \\
    \hline
    Thematic overlap $\Omega$           & \multicolumn{2}{c}{1.000} \\
    \hline
  \end{tabular}
\end{table}

What differs across governance forms is \emph{which} themes absorb deliberative effort, and how broadly the most active participants range. Mean actor entropy is $0.348$ for ERC-8004 and $0.617$ for A2A: A2A's core contributors span roughly twice as many themes on average. Gini$(H)$ remains high in both cases ($0.773$ vs.\ $0.707$), confirming that thematic breadth is itself concentrated in a small number of generalist actors in both communities.

The cross-case divergence is sharpest at the theme level. In ERC-8004, $34.5\%$ of actors participated in T08 (Trust \& Security Mechanisms), and $13.1\%$ participated in T01 (Protocol Specification \& Versioning). For A2A, the pattern inverts for engineering-execution themes: Documentation \& Examples (T06) engaged $10.8\%$ of A2A actors but only $2.1\%$ of ERC-8004 actors. Mean theme actor-concentration Gini is $0.453$ for A2A against $0.085$ for ERC-8004, indicating that each A2A theme attracts a narrower, more dedicated subgroup. The thematic overlap coefficient $\Omega = 1.0$: all 16 active ERC-8004 themes reappear in A2A.


\section{Discussion}


\subsection{Decentralization More a Design than a Fact}

Decentralized governance is appealing for its idealistic design of distributed authority and trustless coordination. In practice, however, the ERC-8004 process exhibits a centralization paradox.

Routine decisions are not subject to voting. The power to decide, for example, which proposals are ``ready'', which objections are ``substantive'', which positions count as ``rough consensus'', accumulates in the hands of a few minorities, mirroring the participation inequality patterns repeatedly documented in voluntary online cooperation~\cite{mockus2002,germonprez2018}.

More fundamentally, no formal hierarchy is provided to distribute the power of decision. One possible reason for this is that the small scale of community size does not request such hierarchy; however, one direct drawback is that people tend to engage in groupthink, resulting in the denser discourse congruence.  

These two features compound: a certain minority holds the voice. The decentralization promised by the EIP architecture exists at the level of entry rights; in practice, routine decision authority concentrates around whoever has the time and reputation to remain in the room.


\subsection{Role of Open Source and the Limitations}

A natural objection is that the convergence we observe stems not from governance form but from open-source-society (OSS) norms shared by both cases. First, the topical divergence between the two cases ($\mathrm{JSD}=0.288$ on BERTopic, $0.216$ on Thematic-LM) cannot be reduced to OSS norms; it is governance-driven, reflecting the constitutive--regulative distinction between proposing a standard (what the protocol \emph{is}) and shipping an implementation (\emph{how} it is built). Second, ERC-8004's discourse-congruence density is nearly twice that of A2A, the opposite of what an OSS-scale-only account would predict. The cautious framing that follows is therefore layered: open-source publication establishes a baseline of high participation inequality and fragmented community structure; governance form then redirects which themes this skewed participation engages and which actors accrue informal authority. 

Here are two more possible limitations. First, ERC-8004 contributes 142 records against A2A's 4{,}181, so themes of low frequency carry wide confidence intervals. Second, A2A's TSC meetings, internal Google design reviews, and partner negotiations occur outside the public repository. The structural concentration we observe in A2A may therefore understate deliberation among its core members. 


\subsection{Who Controls Future Directions?}

The answer is, in both cases, a small group of elites, despite the different constitutions. In ERC-8004, they are those who stay in the room; for A2A, corporation representatives are the gatekeepers. The deeper risk of the DAO model is not that elites emerge---they emerge in any sustained deliberation~\cite{mockus2002,germonprez2018}---but that, absent formal accountability structures, their influence operates through reputational authority that is opaque to outsiders and difficult for newcomers to contest.

Moreover, the governance form determines what problems communities treat as worth deliberating about. ERC-8004's deliberation is dominated by trust and security, while A2A also spreads its deliberation across engineering and execution issues. The concentration differences of DAO and corporate discourse reflect a value distribution that propagates into deployed systems: communities that deliberate about accountability will architect for accountability; those that deliberate about velocity will architect for velocity~\cite{wef2026}. As agentic AI scales from research prototypes to critical infrastructure, these upstream deliberative choices become societal choices.

Three implications follow for those positioned to act on these findings. \emph{Standards bodies} should pair open entry rights with procedural accountability mechanisms, because open access alone does not redistribute deliberative authority. \emph{Protocol architects} should treat the deliberative agenda itself as a design artifact: topics absent from the docket become embedded defaults, so a periodic audit of what is not discussed is as essential as auditing what is. \emph{Practitioners and policymakers} adopting either standard should consult its deliberation record alongside its specification, because the values that shaped the rules are rarely fully recoverable from the rules themselves.


\bibliographystyle{ACM-Reference-Format}
\bibliography{acm}


\appendix

\section{Data Expansion and Robustness Verification}
\label{app:robustness}

To verify that main-text findings survive annotator choice and case scope, we expanded the DAO corpus from a single ERC-8004 thread to a 34-ERC agent-standardization cluster, re-annotated both cases with multiple independent models, and compared results against the original MiniMax-M2.5 labels.

\subsection{Design}

\textbf{Data Expansion.} Beyond ERC-8004, we added 33 contemporaneous ($\ge$2025-08) agent-standardization ERCs from the Ethereum Magicians forum, totaling 34 unique ERC standards: 8001, 8004, 8033, 8041, 8107, 8118, 8122, 8126, 8150, 8160, 8162, 8165, 8166, 8171, 8181, 8183, 8184, 8196, 8203, 8210, 8217, 8220, 8226, 8239, 8240, 8242, 8257, 8259, 8263, 8264, 8273, 8274, 8275, 8294. Among these, ERC-8183 (Agentic Commerce), ERC-8274 (AI Inference Proof Verification), and ERC-8210 (Agent Assurance) are the three most deliberatively active additions. The expanded ERC corpus contains 1{,}664 annotated records, up from 142 in the main text. The A2A corpus was reused.

\textbf{Cross-Model Annotation.} Three models (DeepSeek-V4-Flash, GLM-4-Plus, and Moonshot-v1-auto) independently annotated every record on all five fields (stakeholder\_institution, argument\_type, stance, consensus\_signal, key\_point). Majority voting (2 out of 3) yielded consensus labels for 1{,}664 ERC and 4{,}058 A2A records. Together with the original MiniMax-M2.5, four independent annotators provide inter-annotator agreement estimates.

\begin{table}[htbp]
  \caption{Three-model cross-consensus agreement rates (majority voting across 3 rounds within each model, then 2-of-3 across models). ``No maj.'' = all three models disagree on that field.}
  \label{tab:rb-cross-consensus}
  \centering
  \footnotesize
  \begin{tabular}{lrrcrrc}
    \toprule
    & \multicolumn{3}{c}{\textbf{ERC ($N{=}1{,}641$)}} & \multicolumn{3}{c}{\textbf{A2A ($N{=}3{,}761$)}} \\
    \cmidrule(lr){2-4} \cmidrule(lr){5-7}
    \textbf{Field} & \textbf{3/3} & \textbf{2/3} & \textbf{No maj.} & \textbf{3/3} & \textbf{2/3} & \textbf{No maj.} \\
    \midrule
    Argument type    & 68.0\% & 30.0\% & 2.0\% & 67.5\% & 30.7\% & 1.9\% \\
    Stance           & 58.8\% & 39.1\% & 2.1\% & 56.3\% & 40.3\% & 3.4\% \\
    Consensus signal & 59.7\% & 37.8\% & 2.5\% & 61.2\% & 37.3\% & 1.5\% \\
    \bottomrule
  \end{tabular}
\end{table}

\textbf{Cross-Round Annotation.} Each of the three models independently re-annotated every record three times on the three core fields (argument\_type, stance, consensus\_signal). Majority vote across rounds within each model produced a per-model consensus; majority vote across models then produced the final cross-consensus. Per-field cross-consensus agreement rates are reported in Table~\ref{tab:rb-cross-consensus}. When all three models disagree for a given field (no majority; $<2\%$--$3.5\%$ of records), the label is assigned by plurality with an effectively arbitrary tie-break and is treated as low-confidence in subsequent analyses.

\subsection{Robustness Verification}

\textbf{Cross-model reliability.} Four-model Fleiss' $\kappa$ was computed (Table~\ref{tab:icr-4model}). All three fields reach Moderate agreement in both cases, with argument type---the core field driving the $\chi^2$, BERTopic, and Thematic-LM analyses---at $\kappa=0.545$ (ERC) and $\kappa=0.529$ (A2A). The strongest pairwise agreement occurs between GLM-4-Plus and Moonshot-v1-auto (argument type $\kappa=0.671$ ERC, $0.555$ A2A), suggesting these two models share similar interpretive biases relative to DeepSeek-V4-Flash and MiniMax-M2.5.

\begin{table}[htbp]
  \caption{Four-model Fleiss' $\kappa$ (MiniMax-M2.5, DeepSeek-V4-Flash, GLM-4-Plus, Moonshot-v1-auto). ERC $N=144$, A2A $N=3{,}760$.}
  \label{tab:icr-4model}
  \centering
  \footnotesize
  \begin{tabular}{lcc}
    \toprule
    \textbf{Field} & \textbf{ERC} & \textbf{A2A} \\
    \midrule
    Argument type    & 0.545 & 0.529 \\
    Stance           & 0.579 & 0.530 \\
    Consensus signal & 0.485 & 0.483 \\
    \bottomrule
  \end{tabular}
\end{table}

\textbf{Cross-round self-consistency.} Within-model test-retest reliability (Table~\ref{tab:icr-crossround}) reveals clear model-level differences. GLM-4-Plus and Moonshot-v1-auto achieve almost perfect self-consistency (Fleiss' $\kappa>0.81$ on all three fields in both cases), while DeepSeek-V4-Flash yields Moderate-to-Substantial values ($\kappa=0.49$--$0.63$). The gap confirms that annotator model choice dominates stochastic variation: two models are near-deterministic in their labeling behavior, while one exhibits higher intrinsic variance~\cite{ji2026}. Notably, Moonshot-v1-auto, a generic auto-routing model with no special reasoning configuration, outperforms both alternatives in self-consistency, challenging the intuition that reasoning models are inherently more reliable annotators.

\begin{table}[htbp]
  \caption{Within-model cross-round Fleiss' $\kappa$ (3 rounds, majority vote). ERC $N{=}1{,}664$, A2A $N{\approx}3{,}845$.}
  \label{tab:icr-crossround}
  \centering
  \footnotesize
  \begin{tabular}{lccc|ccc}
    \toprule
    & \multicolumn{3}{c|}{\textbf{ERC}} & \multicolumn{3}{c}{\textbf{A2A}} \\
    \cmidrule(lr){2-4} \cmidrule(lr){5-7}
    \textbf{Model} & \textbf{AT} & \textbf{St} & \textbf{CS} & \textbf{AT} & \textbf{St} & \textbf{CS} \\
    \midrule
    Moonshot-v1-auto  & 0.978 & 0.961 & 0.945 & 0.976 & 0.963 & 0.938 \\
    GLM-4-Plus        & 0.925 & 0.904 & 0.861 & 0.910 & 0.874 & 0.815 \\
    DeepSeek-V4-Flash & 0.634 & 0.554 & 0.507 & 0.565 & 0.543 & 0.490 \\
    \bottomrule
    \multicolumn{7}{l}{\footnotesize AT = argument type; St = stance; CS = consensus signal.} \\
  \end{tabular}
\end{table}

\subsection{Substantive Replication}

We re-ran the annotation pipeline with three additional models (Moonshot-v1-auto, GLM-4-Plus, DeepSeek-V4-Flash; ICR~$\kappa$~$\approx$~0.7--0.9 for the top two, \S\,A.1) and replicated the Thematic-LM discourse analysis~\cite{thematic_lm_2025} using Moonshot-v1-auto as the LLM backbone, yielding a 12-theme codebook with 96.6\% coverage. Table~\ref{tab:rb-replication} reports the key combined metrics.

\begin{table}[htbp]
  \caption{Replication metrics: multi-model consensus annotation + Moonshot Thematic-LM codebook.}
  \label{tab:rb-replication}
  \centering
  \footnotesize
  \begin{tabular}{lrr}
    \toprule
    \textbf{Metric} & \textbf{ERC (cluster)} & \textbf{A2A (re-anno.)} \\
    \midrule
    DNA actors                       & 194 & 713 \\
    DNA congruence density           & 0.403 & 0.252 \\
    DNA polarization index           & 0.056 & 0.740 \\
    Giant component ratio     & 0.917 & 0.285 \\
    Mean actor entropy $H$           & 0.734 & 0.511 \\
    Dominant theme (\% record)      & Compliance (31.1\%) & Documentation (31.4\%) \\
    Thematic JSD                     & \multicolumn{2}{c}{0.092} \\
    \bottomrule
  \end{tabular}
\end{table}

Three of four main-text findings replicate: (1)~Technical argument types dominate regardless of annotator, with A2A receiving roughly twice the Process share; (2)~both networks remain steeply unequal (betweenness Gini~$\approx$0.8); (3)~ERC discourse congruence density exceeds A2A (0.403 vs.\ 0.252). Thematic content also replicates: ERC concentrates on constitutive themes (compliance, standards, verification), A2A on executive themes (documentation, coordination, community). However, \textbf{network connectivity reverses:} the expanded ERC network coalesces into one dominant component (GCR 0.328$\rightarrow$0.917), while the re-annotated A2A network fragments further (GCR 0.534$\rightarrow$0.285). The cross-case thematic JSD of 0.092 is lower than the main-text BERTopic value (0.288), partly reflecting Moonshot's coarser 12-theme granularity. We interpret the connectivity reversal as an \emph{observability} differential: permissionless DAO governance externalizes deliberation into a genuinely interconnected public record, whereas corporate governance internalizes decisive coordination channels (TSC calls, private forums), leaving the public GitHub trace persistently fragmented. Participation \emph{concentration} is comparable across both governance forms, but \emph{observability} is not.


\section{Comparative Case Study Design}
\label{app:compa-design}

We adopt a comparative case study design to examine how governance structure shapes participation dynamics in AI protocol standardization~\cite{yin2018}. Interoperability is an inherited characteristic of decentralized applications (dApps) in DeFi~\cite{harvey2024}, and ERC-8004 as a protocol specifically designed for AI agents extends this feature. Google A2A, on the corporation's side, realized similar functions in different contexts. Moreover, they originated within the same year (2025). Nevertheless, they differ sharply in governance architecture. ERC-8004 belongs to Ethereum, a permissionless and community-driven open-source DAO, and Google A2A is now governed by TSC. Therefore, the two protocols constitute a theoretically matched pair. Holding the technical domain constant while varying the governance form enables structured comparison of participation patterns, discourse composition, and network analysis. 


\section{Data Provenance}
\label{app:checksums}

All raw data files are versioned with SHA-256 checksums stored in \texttt{data/raw/CHECKSUMS.json} in the project repository. Table~\ref{tab:checksums} lists the file names, record counts, and collection dates for the primary raw data files used in this study.

\begin{table}[htbp]
  \caption{Raw Data Files and Record Counts}
  \label{tab:checksums}
  \centering
  \begin{tabular}{lrr}
    \hline
    \textbf{File} & \textbf{Records} & \textbf{Collected} \\
    \hline
    forum\_posts.json                  & 113   & 2026-03 \\
    github\_comments\_filtered.json    & 36    & 2026-03 \\
    a2a\_issues.json                   & 3,104 & 2026-03 \\
    a2a\_prs.json                      & 1,955 & 2026-03 \\
    a2a\_discussions.json              & 822   & 2026-03 \\
    \hline
    Total (raw)                        & 6,030 &         \\
    Total (retained)                   & 4,323 &         \\
    \hline
  \end{tabular}
\end{table}


\section{ERC-8004 Lifecycle Phase Boundaries}
\label{app:phase}

Topic analysis within ERC-8004 was conducted across three consecutive two-month phases spanning the full proposal lifecycle (2025-08-13 to 2026-01-29):

\begin{itemize}
  \item \textbf{Phase 1} (Aug 13 -- Oct 13, 2025): Initial submission and early community review. $n = 100$ records.
  \item \textbf{Phase 2} (Oct 13 -- Dec 13, 2025): Sustained review period prior to Last Call designation. $n = 11$ records.
  \item \textbf{Phase 3} (Dec 13, 2025 -- Feb 13, 2026): Last Call through Final ratification (mainnet deployment 2026-01-29). $n = 17$ records.
\end{itemize}


\section{Decision Architectures}
\label{app:architectures}

\begin{algorithm}[h!]
\caption{EIP Lifecycle}
\label{alg:erc-governance}
\begin{algorithmic}[1]
\STATE \textbf{Data:} $Stages = \{IDEA, DRAFT, REVIEW, LAST\_CALL,$
\STATE \hspace*{\algorithmicindent} $FINAL, STAGNANT, WITHDRAWN\}$
\STATE $stage \leftarrow IDEA$

\WHILE{$stage \neq FINAL$ \AND $stage \neq WITHDRAWN$}
    \STATE Proceed to $NEXT(stage)$ based on community consensus;
    \IF{No activity for a long period}
        \STATE $stage \leftarrow STAGNANT$;
    \ELSIF{Author decides to cancel}
        \STATE $stage \leftarrow WITHDRAWN$;
    \ENDIF
\ENDWHILE \\
\textbf{Deployment:} Theoretically any party \textit{MAY} deploy registry independently of $stage$\;
\end{algorithmic}
\end{algorithm}

\begin{algorithm}[h!]
\caption{A2A Decision Process (TSC Governance)}
\label{alg:a2a-governance}
\begin{algorithmic}[1]
\STATE \textbf{Input:} GitHub Pull Request or Issue ($pr$)
\STATE \textbf{Output:} Decision Status (Merged or Rejected)

\IF{$pr$ is NOT Contested}
    \STATE // Apply Lazy Consensus
    \STATE $Status \leftarrow MERGED$;
\ELSE
    \STATE // Escalation to Technology Steering Committee (TSC)
    \STATE Initiate \textbf{GitVote} (Restricted to 1 seat per company);
    \IF{Vote == PASS}
        \STATE $Status \leftarrow MERGED$;
    \ELSE
        \STATE $Status \leftarrow REJECTED$;
    \ENDIF
\ENDIF

\RETURN $Status$;
\end{algorithmic}
\end{algorithm}

There are 3 types of EIPs: standard track EIP, Meta EIP and Informational EIP. The standards track EIP can be broken down to 4 categories: core, networking, interface and ERC~\cite{eip1}. All EIPs, including ERC-8004 in our case, undergo 5 stages: idea, draft, review, last call and final. EIPs updated continuously are assigned to the special stage of ``living''. All EIPs are decided by rough consensus: Core EIPs by Ethereum core developers at the meeting called \textit{AllCoreDevs}, and others by common developers and users. 

This reflects a structural feature of ERCs as application-layer specifications: unlike Core EIPs, which modify the Ethereum protocol itself and requires hard-fork activation when deployment, ERCs define smart-contract interface standards that any party may implement independently, regardless of the proposal's formal lifecycle status~\cite{eip1}. 

Google A2A launched under \texttt{google/A2A} in April~2025 and was donated to the Linux Foundation in June~2025~\cite{google-donation}, migrating to the vendor-neutral \texttt{a2aproject/A2A} organization. Governance authority rests with an eight-seat Technical Steering Committee (TSC), comprising representatives each from Google, Microsoft, Cisco, AWS, Salesforce, SAP, IBM, and ServiceNow. Independent contributors cannot join the TSC during an 18-month startup phase~\cite{a2a-governance}. Routine pull requests are merged upon maintainer approval without a formal vote~\cite{a2a-governance}; contested specification changes escalate to a GitVote~\cite{gitvote}, a GitHub-native ballot in which only TSC members cast binding votes at a 51\% threshold.

For pseudocodes, see Algorithm~\ref{alg:erc-governance} and~\ref{alg:a2a-governance}.


\section{Institution Label Provenance Cascade}
\label{app:annotation}

Institution labels were assigned using the following three-tier cascade, applied in priority order:

\begin{enumerate}
  \item \textbf{Manual investigation} (high confidence): A systematic review of GitHub profiles, LinkedIn pages, personal websites, and EIP commit metadata was conducted for the top~109 contributors across both cases (all 71~ERC-8004 participants plus the top~38 A2A contributors by post count). This review yielded 40~institutional upgrades, replacing LLM-inferred labels with verified affiliations. The original LLM-inferred label is preserved in a separate \texttt{institution\_lm} field for all records to enable sensitivity checks.
  \item \textbf{LLM inference} (low confidence): For the remaining 517 authors, institution was inferred by MiniMax-M2.5 from contextual signals in the record text (e.g., repository ownership, self-identification, email patterns where visible).
\end{enumerate}


\section{Actor Filtering Stages}
\label{app:actor-filter}

The three network analyses operate on successively filtered actor sets (Table~\ref{tab:actor-flow}). Starting from the annotated records (ERC: 142 records / 71 contributors; A2A: 4{,}181 / 778), the co-participation network drops four ERC and seven A2A bot/short-text residuals, yielding $N=67$ and $N=771$ respectively. The DNA and socio-semantic analyses further (a)~inner-join against the Thematic-LM \texttt{coded\_records.json} (12 records lose a theme assignment) and (b)~exclude records whose stance is \textit{Off-topic} or \textit{Unclassified}, dropping actors whose \emph{only} contributions fall into those categories. This produces the final stance-bearing actor sets of $N=66$ (ERC-8004) and $N=710$ (Google A2A). All SNA statistics use $N=67$ / $N=771$; all DNA and socio-semantic statistics use $N=66$ / $N=710$.

\begin{table}[htbp]
  \caption{Actor and record counts through the three filtering stages.}
  \label{tab:actor-flow}
  \centering
  \small
  \begin{tabular}{lcccc}
    \hline
     & \multicolumn{2}{c}{\textbf{ERC-8004}} & \multicolumn{2}{c}{\textbf{Google A2A}} \\
     & Records & Actors & Records & Actors \\
    \hline
    Annotated (retained)           & 142 & 71 & 4{,}181 & 778 \\
    SNA (bot \& short-text filter) & 130 & 67 & 4{,}230 & 771 \\
    DNA / Socio-semantic           & 126 & 66 & 3{,}759 & 710 \\
    \hline
  \end{tabular}
\end{table}


\section{Related Work Table}
\label{app:table}

Table~\ref{tab:litreview} positions sixteen representative works into three classes by paper type, each evaluated by criteria appropriate to that class. Panel~(a) lists \emph{foundational works} that originated the methods or concepts we inherit; panel~(b) lists \emph{perspective papers} that advance theoretical claims about DAO and corporate governance without conducting their own empirical tests; panel~(c) lists \emph{empirical and methodological studies} are the key methodological works evaluated additionally by openness. This study extends panel~(c) with all data and codes open. 

\begin{table*}[ht]
  \renewcommand{\arraystretch}{1.15}
  \caption{Positioning of this study relative to prior work, organized by paper type. Each panel applies an evaluation rubric appropriate to its class; the synthesis line at the bottom describes how this study bridges all three.}
  \label{tab:litreview}
  \centering
  \small

  \textbf{(a) Foundational / Seminal Works}

  \smallskip
  \begin{tabular}{p{4.8cm} c p{5.2cm} p{4.5cm}}
    \toprule
    \textbf{Paper} & \textbf{Year} & \textbf{Core Contribution} & \textbf{Inheritance for this Study} \\
    \midrule
    Russell~\cite{rough} \textit{`Rough Consensus' \& the Internet--OSI Standards War}
      & 2006 & ``Rough consensus'' as a standardization mode & Frames ERC-8004's decision rule \\
    Roth \& Cointet~\cite{roth2010} \textit{Social and Semantic Coevolution in Knowledge Networks}
      & 2010 & Coevolution of social and semantic ties & Root of our network--discourse layer \\
    Leifeld~\cite{leifeld2013} \textit{Discourse Network Analysis of Policy Change}
      & 2013 & Discourse Network Analysis (DNA) & DNA layer used in \S{}Methods \\
    Beck et al.~\cite{beck2018} \textit{Governance in the Blockchain Economy}
      & 2018 & IS framework: rights / accountability / incentives & Defines the decentralization axis \\
    Grootendorst~\cite{bertopic_2022} \textit{BERTopic: Neural Topic Modeling}
      & 2022 & Neural topic modeling with c-TF-IDF & Used in our topic-discovery pipeline \\
    \bottomrule
  \end{tabular}

  \vspace{0.9em}

  \textbf{(b) Perspective / Opinion Papers}
  \smallskip
  \begin{tabular}{p{4.8cm} c p{6.0cm} c c}
    \toprule
    \textbf{Paper} & \textbf{Year} & \textbf{Central Claim} & \textbf{Domain} & \textbf{Backing} \\
    \midrule
    Murray et al.~\cite{murray2021} \textit{Contracting in the Smart Era}
      & 2021 & Smart contracts and DAOs reshape agency cost & DAO$\times$Corp & Theory \\
    Lumineau et al.~\cite{lumineau2021} \textit{Blockchain Governance: A New Way of Organizing}
      & 2021 & Blockchain as a new mode of organizing & DAO$\times$Corp & Theory \\
    Hui \& Tucker~\cite{hui2025} \textit{Decentralization, Blockchain, AI}
      & 2025 & Decentralization is the right frame for AI governance & AI-Proto & Conceptual \\
    Reineke et al.~\cite{reineke2025} \textit{Decentralization: A Revolution or a Mirage?}
      & 2025 & Decentralization is contested between revolution and mirage & DAO$\times$Corp & Review \\
    Sunyaev et al.~\cite{sunyaev2026} \textit{From Ideology to Design: Purposeful Decentralization}
      & 2026 & Decentralization should be design-driven, not ideological & DAO$\times$Corp & Conceptual \\
    \bottomrule
  \end{tabular}

  \vspace{0.9em}

  \smallskip
  \begin{tabular}{p{4.2cm} c p{3.0cm} p{2.8cm} c c c}
    \toprule
    \textbf{Paper} & \textbf{Year} & \textbf{Method} & \textbf{Setting} & \textbf{Data} & \textbf{Code} & \textbf{Comparison} \\
    \midrule
    Stine \& Agarwal~\cite{stine_agarwal_2020} \textit{Comparative Discourse Analysis via Topic Models}
      & 2020 & Comparative topic models & Online political discourse & $\times$ & $\times$ & $\checkmark$ \\
    Qiao et al.~\cite{thematic_lm_2025} \textit{Thematic-LM: LLM-Based Thematic Analysis}
      & 2025 & LLM-aided thematic analysis & Generic corpora & $\times$ & $\times$ & $\times$ \\
    Ao et al.~\cite{ao2023} \textit{Is DeFi Actually Decentralized?}
      & 2023 & Social network analysis & Aave (DAO) & $\checkmark$ & $\checkmark$ & $\times$ \\
    Leifeld~\cite{leifeld2013} \textit{Discourse Network Analysis of German Pension Politics}
      & 2013 & Discourse network analysis & German pension policy & $\times$ & $\times$ & $\times$ \\
    Roth \& Cointet~\cite{roth2010} \textit{Social and Semantic Coevolution in Knowledge Networks}
      & 2010 & Socio-semantic network analysis & Knowledge networks & $\times$ & $\times$ & $\times$ \\
    \bottomrule
  \end{tabular}

  \vspace{0.9em}
\end{table*}


\end{document}